\documentclass{article}

     \usepackage[preprint, nonatbib]{neurips_2022}

\usepackage{times}
\usepackage{epsfig}
\usepackage{graphicx}
\usepackage{amsmath}
\usepackage{amssymb}
\usepackage{balance}
\usepackage{flushend}

\usepackage{color}
\usepackage[utf8]{inputenc}
\usepackage{multirow}
\usepackage{array}
\usepackage{adjustbox}
\usepackage{bbding}
\usepackage{makecell}

\usepackage{pifont}

\usepackage[utf8]{inputenc} 
\usepackage[T1]{fontenc}    
\usepackage{hyperref}       
\usepackage{url}            
\usepackage{booktabs}       
\usepackage{amsfonts}       
\usepackage{nicefrac}       
\usepackage{microtype}      
\usepackage{xcolor}         

\title{Emergence of Double-slit Interference by Representing Visual Space in Artificial Neural Networks}

\author{%
  Xiuxiu Bai\thanks{Corresponding author}, Zhe Liu, Yao Gao, Bin Liu, Yongqiang Hao\\
  School of Software Engineering\\
  Xi'an Jiaotong University\\
  \texttt{xiubai@xjtu.edu.cn} \\
  \texttt{ \{alfredliu, yaogao, csliubin, hyq123\}@stu.xjtu.edu.cn}\\
}

\begin{document}

\maketitle

\begin{abstract}
Artificial neural networks have realized incredible successes at image recognition, but the underlying mechanism of visual space representation remains a huge mystery. Grid cells (2014 Nobel Prize) in the entorhinal cortex support a periodic representation as a metric for coding space. Here, we develop a self-supervised convolutional neural network to perform visual space location, leading to the emergence of single-slit diffraction and double-slit interference patterns of waves. Our discoveries reveal the nature of CNN encoding visual space to a certain extent. CNN is no longer a black box in terms of visual spatial encoding, it is interpretable. Our findings indicate that the periodicity property of waves provides a space metric, suggesting a general role of spatial coordinate frame in artificial neural networks. 
\end{abstract}

\section{Introduction}

The ability to localize in visual space is crucial in visual representation and recognition. Grid cells (2014 Nobel Prize)  in entorhinal cortex support a periodic representation as a metric for coding navigational space\cite{Krupic2015GridCS, Stensola2012TheEG, Hafting2005MicrostructureOA, Mathis2012OptimalPC, Bush2015UsingGC}. The underlying mechanism of periodic firing may be attributed to periodic band waves derived from a series of wavelengths and orientations\cite{Krupic2012NeuralRO}. Recent evidence reveals that the human entorhinal cortex represents locations in visual space with a grid code \cite{Nau2018HexadirectionalCO, Julian2018HumanEC}. Artificial neural networks have proved incredible successes at image recognition. However, it remains unclear how artificial networks encode visual space or whether such a grid-like coding exists.

From the perspective of visual neuroscience, human visual systems establish a coordinate frame system, which dominates the concept of mental space in the process of image recognition. In the physical layer, space and object features are encoded by two relatively independent visual pathways, dorsal flow, and ventral flow. 

From the perspective of artificial neural networks, all features including image space and object are hard-coded in models, resulting in requiring a massive collection of samples to learn data distributions with various small changes. Moreover, the existing models lack a coordinate frame and explicit space metric concept. This is critical for the generalization ability of current networks beyond training distribution.

As a novel way to understand neural signals, some studies trained recurrent neural networks to conduct navigation tasks based on path integration, resulting in grid-like patterns emerging in models \cite{Banino2018VectorbasedNU, Cueva2018EmergenceOG, Gao2019LearningGC}. This evidence supports that neural responses in the entorhinal cortex possess grid code to map path-based self-locations. The previous spatial navigation tasks focus on simulating the process of path integration to observe the patterns of rate maps. If there is no similar path integration mechanism, how will visual space be represented in artificial neural networks? However, it remains a large mystery.

The critical difference between visual space representation and navigation tasks is as follows. The spatial navigation task relies on path integration to encode maps, which can be conducted without visual cues. The required inputs are speed and direction, and the output is trajectory. In visual space representation, the input is an image with objects, and the output is the spatial position coding of objects.

We have addressed this issue by introducing a self-supervised convolutional neural network (CNN) to explore the nature of visual space representation. We train a convolutional neural network to perform visual space location in an image. The network is required to learn its self-location in a self-supervised manner. We expose the emergence of single-slit diffraction and double-slit interference patterns of waves in the trained networks. Our findings indicate that the periodicity property of waves provides a space metric, suggesting a general role of coordinate frame in encoding signals in visual space.

The contribution of our paper is as follows.

\begin{itemize}

\item[$\bullet$] We discover wave patterns in convolutional neural networks learning visual space representation. Our findings reveal the nature of CNN encoding visual space to a certain extent. CNN is no longer a black box in terms of visual spatial encoding, it is interpretable. 

\item[$\bullet$] Our findings will inspire researchers to explore the robustness of CNN from a new perspective of the wave.

\end{itemize}

\section{Related work}

This section will introduce and analyze the related studies from two perspectives of neuroscience and artificial neural networks.

\textbf{Neuroscience.} Grid Cells in the entorhinal cortex have the hexagonal firing structure, which exhibits a map of an animal’s position in an open environment \cite{Krupic2015GridCS, Stensola2012TheEG, Hafting2005MicrostructureOA, Mathis2012OptimalPC}. This regular pattern of grid coding and its spatial periodicity provides an interpretable metric of space representation \cite{Hafting2005MicrostructureOA}. Grid cells are considered to conduct path integration based on self-motion cues during navigation \cite{Bush2015UsingGC}. 

Beyond navigation, recent studies provide evidence for grid-like cells for representing visual space, suggesting that visual grid codes may utilize eye movements to update allocentric gaze location \cite{Nau2018HexadirectionalCO, Julian2018HumanEC}. In these cognitive experiments, as participants watched a series of images, grid coding patterns emerged in the cerebral cortex, to achieve visual spatial representation.

Moreover, in the visual cortex, the input information is roughly decoupled into two separate pathways. The dorsal stream dominates “where” signals, and the ventral stream dominates “what” signals.

\textbf{Artificial neural networks.} Some studies trained recurrent neural networks to conduct navigation tasks based on path integration, resulting in grid-like patterns emerging in learned models \cite{Banino2018VectorbasedNU, Cueva2018EmergenceOG, Gao2019LearningGC}. These pieces of evidence support that neural responses in the entorhinal cortex possess grid code to map path-based self-locations.

The previous spatial navigation tasks focus on simulating the process of path integration to observe the patterns of rate maps. If there is no similar path integration mechanism, how will the visual space representation be solved by artificial neural networks? However, it remains unclear.

Inspired by separate pathways in the visual cortex, Ebrahimpour et al. \cite{Ebrahimpour2019VentralDorsalNN} proposed a ventral-dorsal network for detecting objects. The Ventral Net removes backgrounds using a selective activation cue. The Dorsal Net recognizes objects of interest in reserved areas. It provides a faster focus to the detection. However, visual space coding rules of artificial neural networks remain elusive.

In existing space-related tasks, the network is forced to self-supervised learn the feature extraction, and it then serves as a pretext task to replace the supervised pre-trained model \cite{Noroozi2016UnsupervisedLO, Gidaris2018UnsupervisedRL, Caron2020UnsupervisedLO}. For instance, in the jigsaw puzzle task \cite{Noroozi2016UnsupervisedLO}, the model relies on the appearance and shape information to learn the spatial arrangement of parts. It leads to the final hard-coding of space and object intrinsic features in trained models, and it is difficult to observe spatial coding patterns.

In this paper, we focus on visual space representation. We train a convolutional neural network to perform visual space location. The network is required to learn its self-location in a self-supervised manner.


\begin{figure*}[t]
\begin{center}
\includegraphics[width = 0.8 \linewidth]{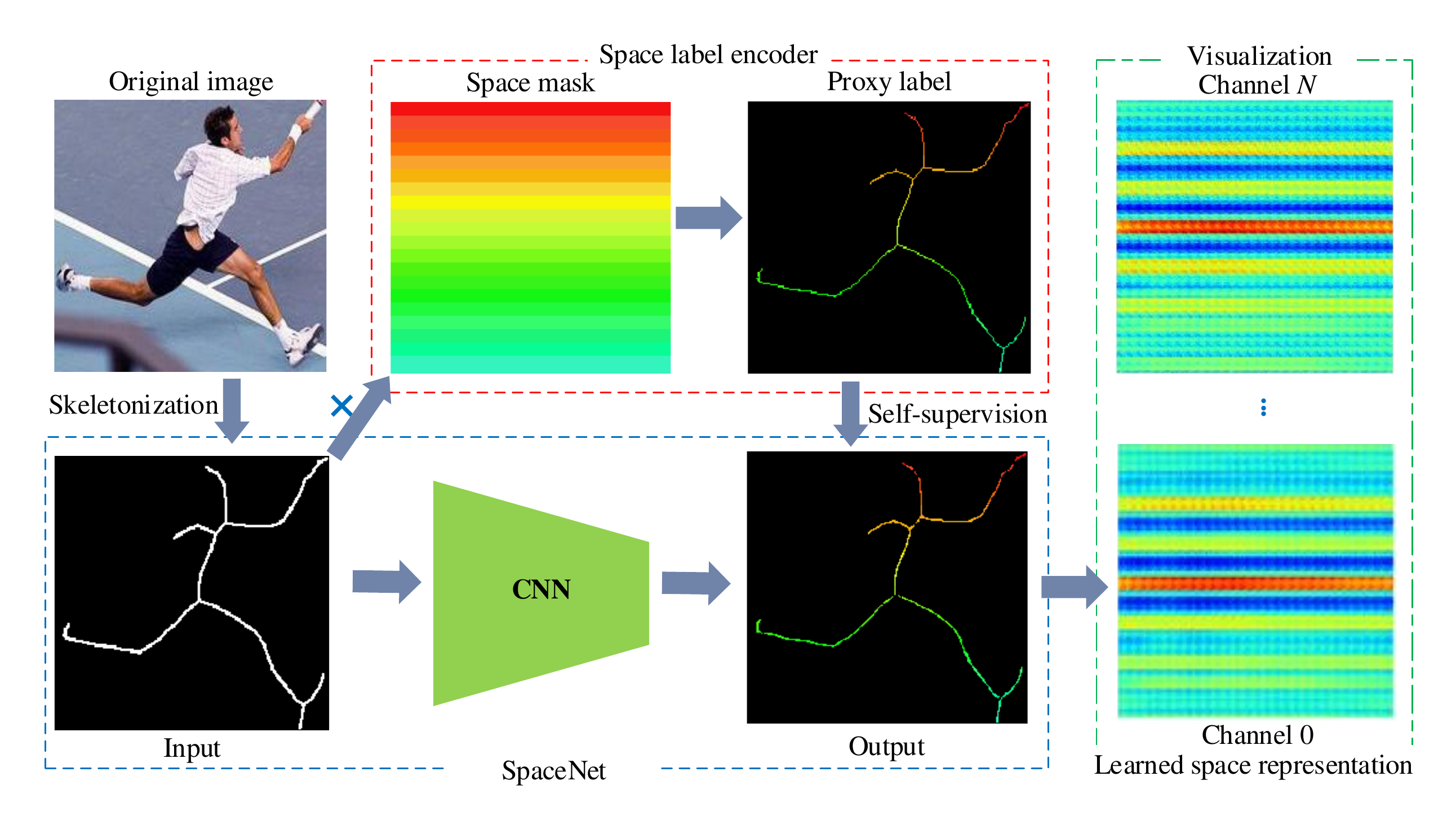}
\end{center}
\caption{SpaceNet architecture. The input is the object skeleton. The proxy label is the skeleton with spatial position information. The loss uses the weighted cross entropy loss. The input without appearance information can force the model to solely learn spatial representation patterns. The learned visual space representation exhibits the periodic property of waves.}
\label{fig1}
\end{figure*}

\begin{figure}[t]
\begin{center}
\includegraphics[width = 0.55  \linewidth]{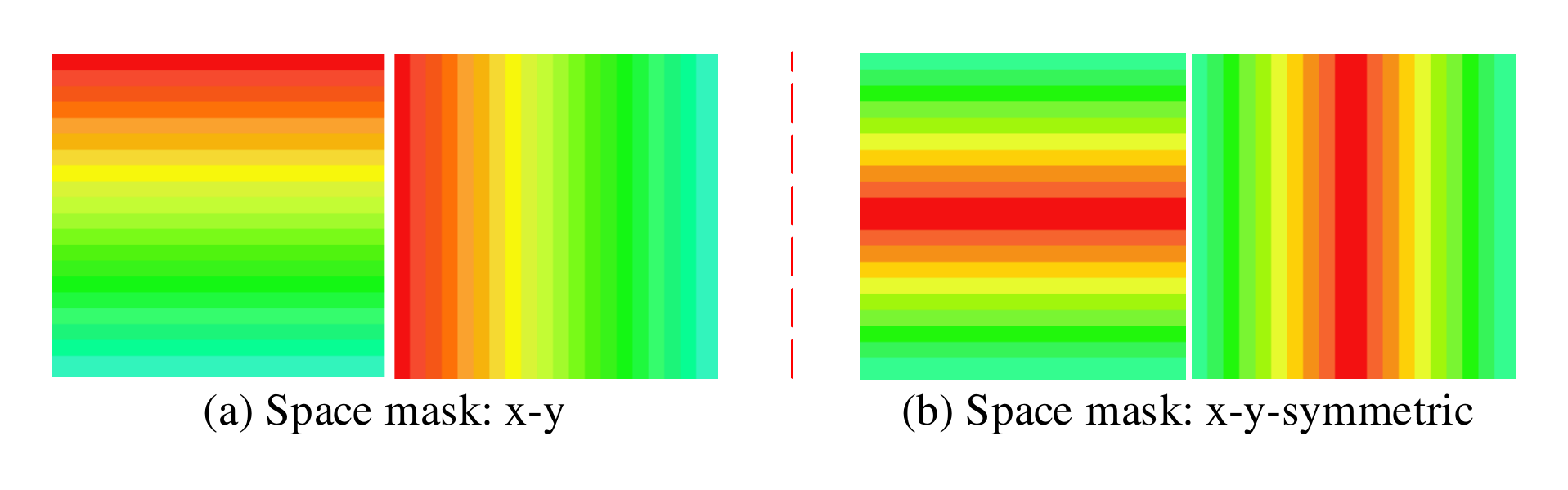}
\end{center}
\centering
\caption{Space masks of proxy labels.}
\vspace{-1em}
\label{fig2}
\end{figure}


\section{Methods}

To observe spatial coding patterns in learned models, we introduce a skeleton-based self-supervised visual space representation model, namely SpaceNet. The core idea is to adopt the full convolution neural network to learn visual spatial coding in a self-supervised manner. In SpaceNet, the object skeleton is used as input, and the corresponding spatial position code of the skeleton is used as the proxy label, which forces the target model to learn the visual spatial coding ability. 
The advantage of using a skeleton image as input is that the input has no relevant texture or color information, so there is no shortcut for the network, which can force models to learn spatial features to reduce the loss.

Figure~\ref{fig1} shows the SpaceNet architecture. First, we construct a pair of input and proxy labels. Our central goal is to observe how spatial features are encoded. To avoid the interference of image textures and colors, we remove the appearance information from the natural image and only retain object skeletons as inputs. The proxy label is the skeleton with spatial position information. It can be obtained by multiplying the skeleton image and a pair of orthogonal space masks. The loss uses the cross entropy loss function.

We design two kinds of spatial masks of proxy labels to train networks, as shown in Figure~\ref{fig2}. 

1) x-y space mask: The space mask divides evenly into horizontal and vertical directions, and encodes in ascending order. 

2) x-y-symmetric space mask: The space mask divides evenly into horizontal and vertical directions, and symmetrically encodes from the middle to both sides.

The space mask divides evenly in horizontal and vertical directions, and encodes in ascending order. Denoting the input and proxy label as $(x, y)$, the space mask $M_{i}$ is as follows:
\begin{equation}
y = x \times M_{i}  
\end{equation}
\begin{equation}
M_{i,j} = k, i \in \left \{0,1\right \}, j\in [0,N)
\end{equation}
where $i$ represents directions, $i = 0$ is horizontal direction, $i =1$ is vertical direction. $N$ denotes the number of regions in the space mask. $k$ represents space location and $k$ is a integer. For x-y space mask, $k\in [1,N]$; For x-y-symmetric space mask, $k\in [1,N/2]$. In the experiments, we use $N = 20$.

After the (input, proxy label) set is generated, it can force the SpaceNet to self-supervised learning visual space representation. 
 The network architecture of SpaceNet is based on FCN8 \cite{Shelhamer2017FullyCN}. The used backbone model is VGG16 \cite{Simonyan2015VeryDC}.

The loss uses the weighted cross entropy loss.
We use the weight to balance between objects and backgrounds.
 The balance weight $w(m)$ is as follows:
\begin{equation}
	w(m)= \left\{\begin{matrix}
	\frac{O^{(n)}}{X^{(n)}} &,X^{(m)} = 0 \\ 
	 \frac{B^{(n)}}{X^{(n)}} & ,X^{(m)} > 0
	 \end{matrix}\right.
\end{equation}
where $X^{(n)}$, $O^{(n)}$ and $B^{(n)}$ represent the number of image, object and background pixels. $m$ denotes each pixel.

\begin{figure*}[t]
\begin{center}
\includegraphics[width = 0.7 \linewidth]{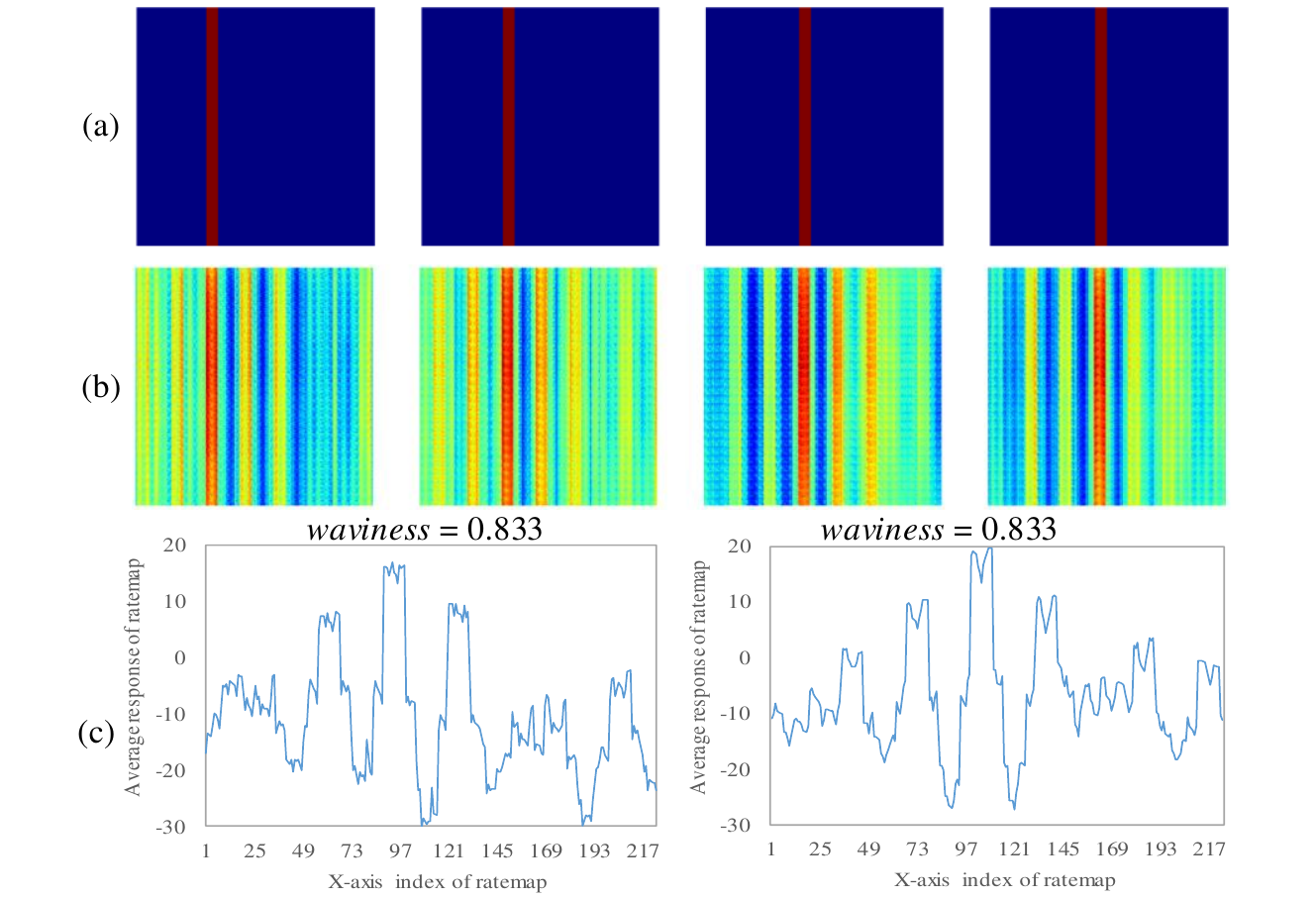}
\end{center}
\caption{Single-slit diffraction patterns of waves emerge in a network trained to represent visual space. (a) Proxy label: one-hot encoding. (b) After training: the last hidden layer exhibits spatially responses resembling single-slit diffraction patterns of waves. (c) Quantitative results of single-slit diffraction patterns.}
\label{fig3}
\vspace{0em}
\end{figure*}

\begin{figure*}[t]
\begin{center}
\includegraphics[width = 0.7 \linewidth]{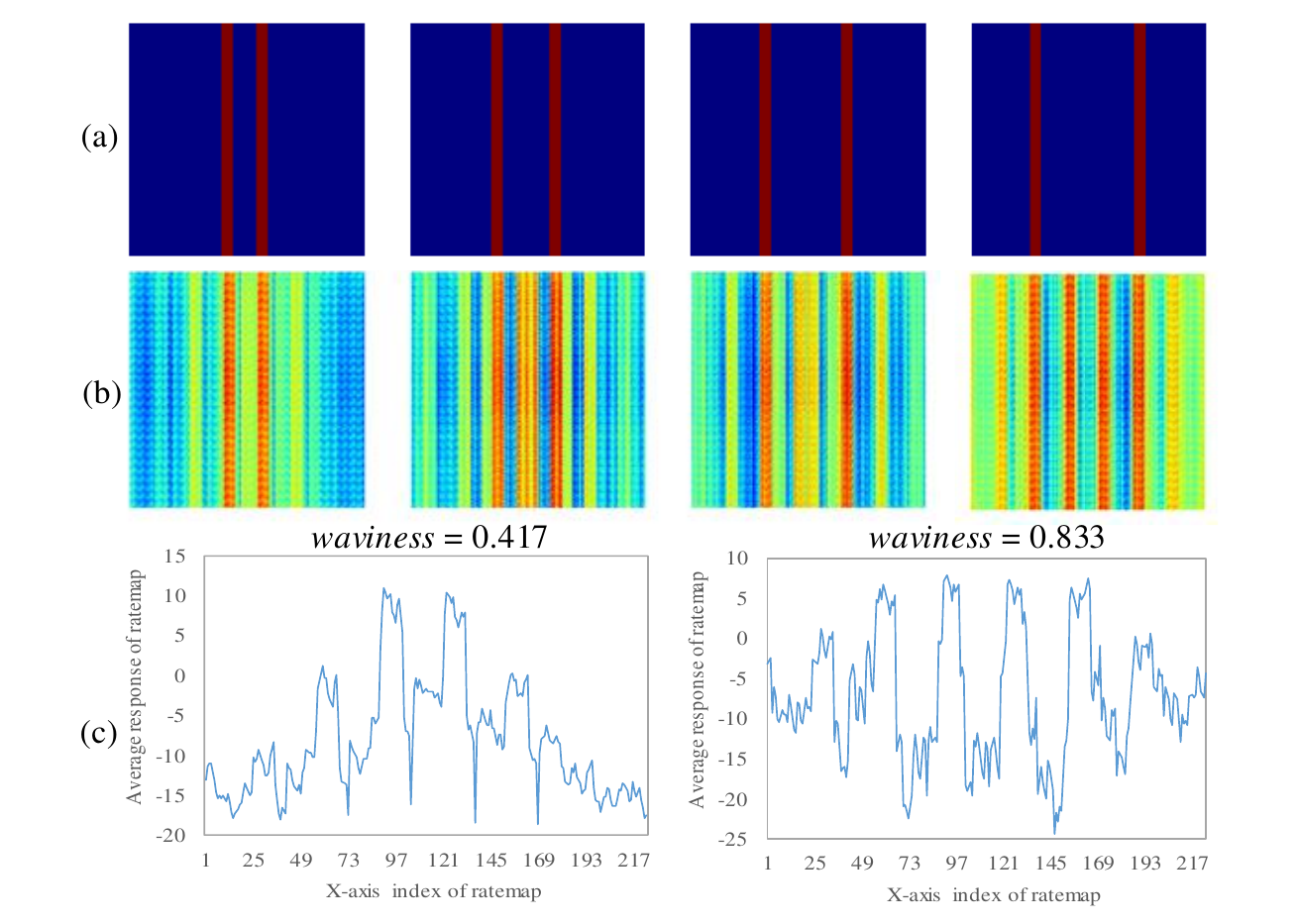}
\end{center}
\caption{Double-slit interference patterns of waves emerge in a network trained to represent visual space. (a) Proxy label: one-hot encoding. (b) After training: the last hidden layer exhibits spatially responses resembling double-slit interference patterns of waves. (c) Quantitative results of double-slit interference patterns.}
\label{fig4}
\end{figure*}

 \section{Experiments}
 
 We develop a self-supervised convolutional neural network to explore the nature of visual space representation.
 
\subsection{Implantation details}
 
 The architecture of SpaceNet is based on FCN8 \cite{Shelhamer2017FullyCN}. The used backbone model is VGG16 \cite{Simonyan2015VeryDC}. 
We use skeleton datasets SK-LARGE\cite{shen2017deepskeleton} and SYM-PASCAL\cite{Ke2017SRN} to train SpaceNet. The training set is ground-truth labels of skeleton training datasets. The testing set is ground-truth labels of skeleton testing datasets. The input images are resized to 224 $\times$ 224.

The SK-LARGE dataset contains 746 training images and 745 testing images. The SYM-PASCAL dataset consists of 647 training images and 745 testing images. During training, we use data augmentation strategies including rotating input images to four directions $(0^\circ, 90^\circ, 180^\circ, 270^\circ)$.

We use the PyTorch framework to implement. During training, we use an SGD optimizer. Momentum is set to 0.9 and weight decay is set to 0. We train the network for 80 epochs. The initial learning rate is 0.1 and it is reduced by 10 times after every 30 epochs. The batch size is set to 16. We conduct all training stages on a single NVIDIA GeForce RTX 2080Ti GPU.

\subsection{Metric}
We define a $waviness$ to measure the probability of wave patterns.
When significant peaks and troughs alternately appear in a distribution, it can be considered a wave pattern. 
1) If the local difference $Ab_{diff}(n)$ divided by global difference $A_{diff}$ is greater than a threshold $T$, there is a region that may be a crest or trough. 
2) The sign of two adjacent local difference values needs to be opposite, to ensure that the two adjacent effective intervals are alternating peaks and troughs. 
3) $waviness$ equals the sum of effective wavelengths $\lambda_{n}$ is divided by the length of global distribution $L$.
\begin{equation}
\begin{aligned}
&waviness = 	\sum_{n=1}^{N-2}\lambda_{n} / L,  s.t. \left| \frac{Ab_{diff}(n)} {A_{diff}} \right| =\delta>T, \\
&sign(Ab_{diff}(n))+sign(Ab_{diff}(n+1))=0 \\
\end{aligned}
\end{equation}
where $waviness = 0$ indicates the distribution has no wave pattern at all. $waviness = 1$ indicates that the distribution has strong wave patterns, such as standard square wave or sine wave.  $waviness \in [0,1]$. 
In our experiments, the threshold $T$ is 0.1.

We use weighted pixel-level classification accuracy as spatial position prediction accuracy.
The purpose of weighting is that the object accounts for a small proportion of the total image. If it is not weighted, it will lead to high prediction accuracy, but the actual qualitative results are poor. 
The used spatial position prediction accuracy $Acc_{space}$ is as follows.
\begin{equation}
\begin{aligned}
& Acc_{space} = \frac{B^{(n)}}{X^{(n)}} Acc_{o} + \frac{O^{(n)}}{X^{(n)}}Acc_{b} \\
\end{aligned}
\end{equation}
where $X^{(n)}$, $O^{(n)}$ and $B^{(n)}$ represent the number of image, object and background pixels.

\subsection{How to obtain ratemaps?}
Each ratemap is obtained as follows. It tests all testing images of each dataset to get feature maps. Then, it calculates mean values over all features maps with the same channel on the same hidden layer, which are ratemaps. Why do we use the mean values of feature maps over the whole test set as the ratemap? It is because the feature map of a single testing image contains the individual activated object information besides space features. The mean activity of the whole test set can eliminate the activated object information.

\begin{figure*}[t]
\begin{center}
\includegraphics[width = 1 \linewidth]{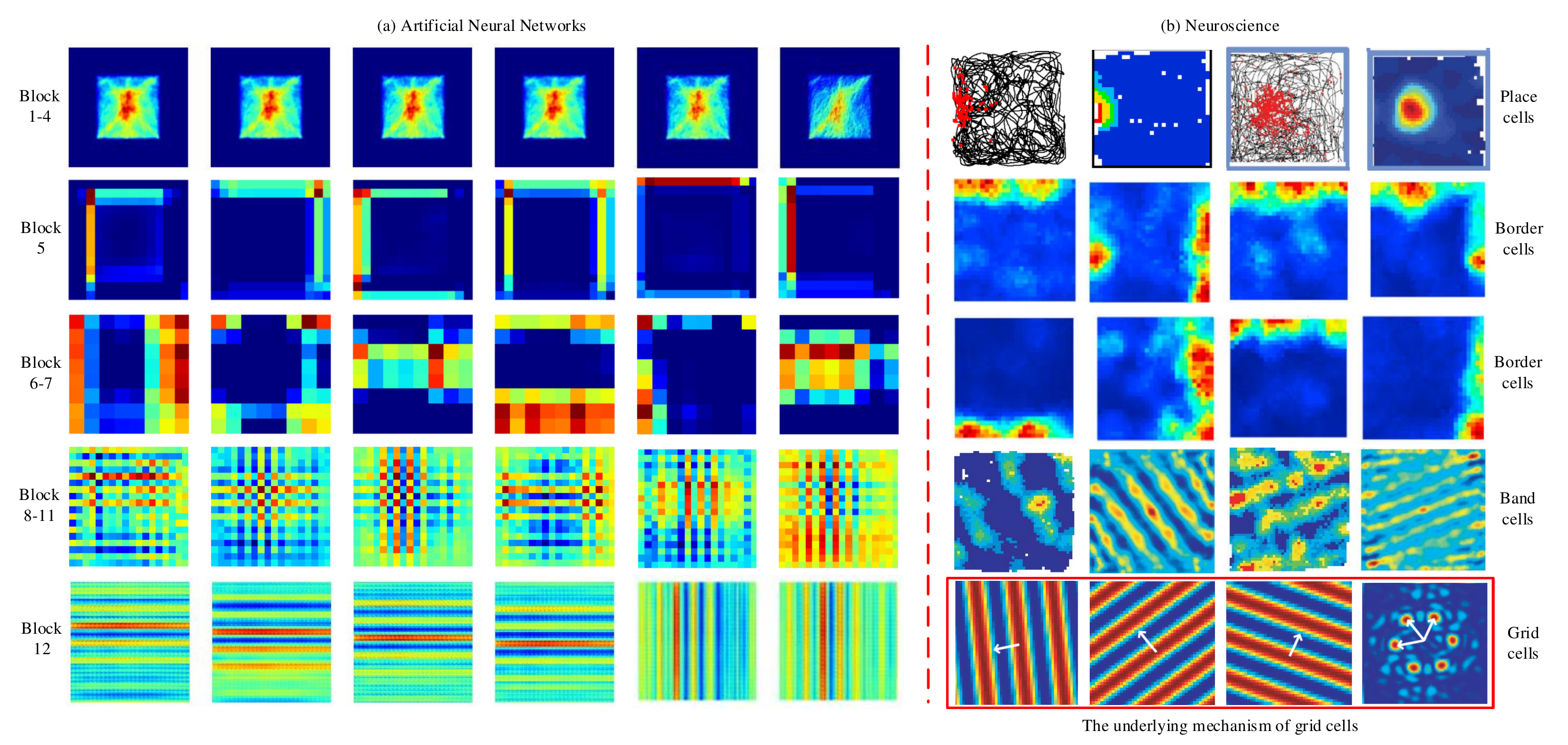}
\end{center}
\caption{Spatial ratemaps from different block layers in trained networks to represent visual space.  From shallow to deep layers of trained networks, units display spatial response patterns resembling place cells \cite{Barry2014NeuralMO, moser2015place}, border cells \cite{Bjerknes2014RepresentationOG} and spatially periodic band cells \cite{Krupic2012NeuralRO} in neuroscience.
Place cells are those that fire when the animal enters a specific location.
Border cells are those that fire when the animal approaches one or some local geometric boundaries.
Band cells indicate the animal firing in a spatially periodic band of locations across the environment.
Grid cells indicate the animal firing in a hexagonal array of locations across the environment.
Spatially periodic firing patterns (grid cells and band cells) consist of plane waves extracted from a set of orientations and wavelengths \cite{Krupic2012NeuralRO}. 
}
\label{fig5}
\end{figure*}

\subsection{Analyses of network units}
The direct goal of SpaceNet is to locate the positions of object skeletons. The underlying aim of SpaceNet is to observe the coding patterns of spatial features in models.

 After training, the location accuracy of all the trained models can reach above 90\%. More importantly, we further analyze ratemaps of network units. In our paper, we focus on how neural networks learn to represent space.

The possible spatial ratemaps of trained models are as follows. 

1) The target locations are activated and other non-targeted locations are inhibited.

2) The target locations are activated and the activated probability of other locations decreases with distances from target locations. It resembles Gaussian distribution.

3) The target locations are activated and other non-targeted locations are irregularly activated.

4) The target locations are activated and other non-targeted locations are regularly activated.

We observe ratemaps of the last hidden layer of models. What surprised us is that the distribution of ratemaps exhibits periodic patterns of waves. When the target location is a single slit, the diffraction patterns emerge on spatial activity maps (Figure~\ref{fig3}). When the target location is double slits, the interference patterns emerge on spatial activity maps (Figure~\ref{fig4}). 

We find that diffraction and interference patterns appear in the visual spatial coding based on artificial neural networks, which verify it possesses the periodicity of waves.
It is based on the wave theory, that is, a pattern is considered to be a wave if diffraction or interference occurs. Therefore, we can use the wavelength of this visual space coding pattern to measure visual space.

How does the network form such surprisingly spatially periodic waves? We report spatial ratemaps from different block layers in networks to represent visual space (Figure~\ref{fig5}). The activity maps of the first four block layers are directly related to the egocentric (object-centered) object, and the activated locations are distributions of objects themselves. The activity maps of the back block layers are related to allocentric (world-centered) representations of spatial locations. The spatial features have been decoupled from object features. From shallow to deep layers of trained networks, activity maps exhibit spatial response patterns resembling place cells \cite{Barry2014NeuralMO}, border cells \cite{Bjerknes2014RepresentationOG} and stable periodic band cells\cite{Krupic2012NeuralRO}. This demonstrates the emergence of periodic waves in trained networks to represent visual space without path integration.

In addition, the underlying mechanism of gird firing may be attributed to periodic bands derived from a series of wavelengths and orientations\cite{Krupic2012NeuralRO}. Our results support the theory that the formation of gird cells may be attributed to spatially periodic bands.

 \subsection{Ablation study}
We conduct the ablation study to verify the universality of wave patterns in trained CNN models encoding visual space.
Table~\ref{table1} shows the influence of various factors on the emergence of wave patterns in SpaceNet. 
The same wave pattern appears in different network architectures, backbones, initialization methods, shapes of input images, shapes of space masks, and datasets.
The spatial representation of the shallow network has low accuracy and no obvious wave pattern in the ratemaps.

 \begin{table*}[t]
\centering
\caption{The influence of various factors on the emergence of wave pattern in SpaceNet. $\surd$ means it can obviously observe wave patterns.}
\label{table1}
\renewcommand\arraystretch{1.5}
\begin{adjustbox}{width= 0.7 \linewidth}
\scriptsize
\begin{tabular}{l|l|c|c|c}
\hline
Factors & Types & $Acc_{space}$ $\uparrow$ & $waviness$ $\uparrow$ & wave pattern \\
\hline
\multirow{2}{*}{Architecture } & FCN8s & 0.983 & 0.581 & $\surd$ \\
\cline{2-5}
&DeepLabv2  & 0.968 & 0.695 & $\surd$\\
\cline{1-5}
\hline
\multirow{2}{*}{Backbone} & VGG16 & 0.983 & 0.581 & $\surd$ \\
\cline{2-5}
&ResNet50 & 0.941 & 0.700 & $\surd$ \\
\cline{1-5}
\multirow{2}{*}{Initialization} & Random & 0.983 & 0.581 & $\surd$\\
\cline{2-5}
&Pretrained ImageNet&  0.954 & 0.696 & $\surd$\\
\cline{1-5}
\multirow{2}{*}{Shape of inputs} & Square & 0.983 & 0. 581 & $\surd$\\
\cline{2-5}
&Non-square& 0.923 & 0.369 & $\surd$\\
\cline{1-5}
\multirow{2}{*}{Shape of masks} & x-y & 0.983 & 0. 581 & $\surd$\\
\cline{2-5}
&x-y-symmetric & 0.983 & 0.550 & $\surd$\\
\cline{1-5}
\multirow{2}{*}{Datasets} & SK-LARGE & 0.983 & 0.581 & $\surd$\\
\cline{2-5}
&SYM-PASCAL & 0.982 & 0.581 & $\surd$\\
\cline{1-5}
Layers &Shallow (3 layers)& 0.393 & 0.102 & {\Large $\times$} \\
\hline
\end{tabular}
\end{adjustbox}
\end{table*}

\section{Discussion: The value of discovery}
\textbf{Experimental phenomena:}
1) In the sole spatial representation task, wave patterns emerge in deep layers of CNN. 

2) In the spatial-related tasks, no wave pattern can be observed in deep layers. Spatial-related tasks include classification, semantic segmentation, object detection, skeleton detection, and so on. These tasks need to implicitly or explicitly encode space to locate and recognize.

Why is it difficult to directly observe the wave patterns in the feature maps of these spatial-related models? The main reason is that these CNN models encode object and spatial features together, so it is difficult to observe sole spatial-related coding patterns. In our SpaceNet, we only learn the visual space representation task, and remove backgrounds, object color and appearance features in the input image, which can force the model to solely learn spatial representation patterns. So this amazing spatial encoding pattern can be founded.

In spatial-related tasks, is there a periodic wave pattern in the coding of spatial features? How to verify it?

\textbf{Hypothesis:}
We use waves to attack CNN models of these spatial-related tasks, and test whether the CNN model is sensitive to the wave pattern and whether it is sensitive to the wavelength and orientation of waves. If it can attack the CNN model effectively, we can indirectly verify the existence of wave patterns in the receptive field of CNN models.

\textbf{Verifying hypothesis:}
We will use waves to attack spatial-related models such as classification, semantic segmentation, object detection, skeleton detection, and so on. Classification and skeleton detection need to implicitly locate objects in space. Semantic segmentation and object detection need to explicitly locate objects in space.

1)	Classification. Experiments in \cite{Guo2019SimpleBA} verify that a group of waves can effectively attack classification models. 

2)	Skeleton detection. Experiments in \cite{Bai2020OnTR} verify that the simplest square wave has a strong attack effect on skeleton detection models, and skeleton detection models are very sensitive to the wavelength and orientation of the added square wave noises. 

3)	Semantic segmentation. We use the simple square wave in \cite{Bai2020OnTR} to attack the object detection model, and the attack effect is significant. The used model is DeepLabv2 \cite{Chen2018DeepLabSI}.

4)	Object detection. We use the simple square wave in \cite{Bai2020OnTR} to attack the object detection model, and the attack effect is significant. The used model is YOLOv4 \cite{Bochkovskiy2020YOLOv4OS}.

\begin{figure*}[]
\begin{center}
\includegraphics[width = 0.98 \linewidth]{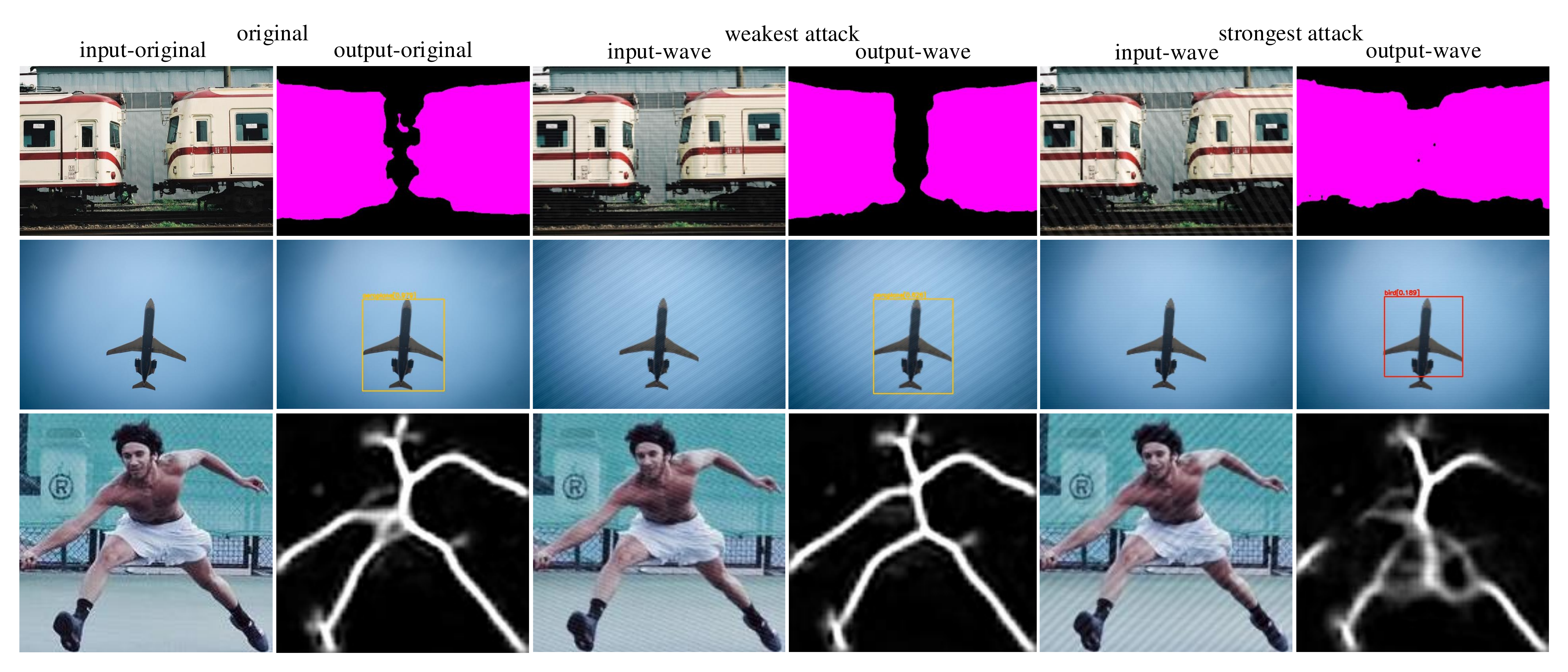}
\end{center}
\caption{Spatial-related models are sensitive to wavelength and orientation of waves. 
Row 1 is the result of semantic segmentation model DeepLabv2-ResNet101\cite{Chen2018DeepLabSI}. 
Row 2 is the result of object detection model YOLOv4-CSPDarknet53 \cite{Bochkovskiy2020YOLOv4OS}. 
Row 3 is the result of skeleton detection model ProMask-VGG16 \cite{Bai2020ProMaskPM}.
Columns 1 and 2 are original inputs and outputs. 
Columns 3 and 4 are inputs and outputs of the weakest wave attack. 
Columns 5 and 6 are inputs and outputs of the strongest wave attack. The attack intensity is $\epsilon = 8$.
}
\label{fig6}
\vspace{0em}
\end{figure*}

Table~\ref{table2} shows the quantitative comparison results of space-related models when they are attacked by waves. 
Figure~\ref{fig6} shows the qualitative comparison results of space-related models when they are attacked by waves.
The used attack method is a black-box Frequency attack \cite{Bai2020OnTR}, which searches the optimization wavelength and orientation of square waves to disrupt predicting results. The wave with a specific wavelength and orientation is added to the original image as adversarial samples. 
Spatial-related models are sensitive to the wavelength and orientation of waves. It can search for the strongest attack wave to disrupt predicting results. Conversely, it can find the weakest attack wave to improve predicting results.

 \begin{table*}[t]
\centering
\caption{Wave attack on space-related tasks. The used attack method is a black-box Frequency attack \cite{Bai2020OnTR}, which searches the optimization wavelength and orientation of square waves to disrupt predicting results. The attack intensity is $\epsilon = 8$. 
The strongest attack is to search for a specific wavelength and orientation, so that the disruption ability is the strongest.
The weakest attack is to search for a specific wavelength and orientation, so that the attack ability is the weakest.
}
\label{table2}
\renewcommand\arraystretch{1.8}
\begin{adjustbox}{width= 0.98 \linewidth}
\resizebox{\textwidth}{!}{
\begin{tabular}{l|l|l|l|l|l|l|l}
\hline
\multirow{3}{*}{Task}&\multirow{3}{*}{Model}&\multirow{3}{*}{Dataset}&\multirow{3}{*}{Metric}&\multicolumn{4}{c}{Performance comparison}\\
\cline{5-8}
&&&&\multirow{2}{*}{Original}&\multirow{2}{*}{\makecell[l]{Random\\Gaussian\\noise}}&\multicolumn{2}{c}{Wave attack}\\
\cline{7-8}
&&&&&&\makecell[l]{Weakset\\attack}&\makecell[l]{Strongest\\attack}\\
\hline
\makecell[l]{Semantic\\segmentation}&\makecell[l]{DeepLabv2-\\ResNet101\cite{Chen2018DeepLabSI}}&VOC-val \cite{Everingham2014ThePV} &MIOU $\uparrow$&0.765&0.754& \textbf{0.799} &\textbf{0.501}\\
\hline
\makecell[l]{Object\\ detection}&\makecell[l]{YOLOv4-\\CSPDarknet53 \cite{Bochkovskiy2020YOLOv4OS}}&COCO-val \cite{Lin2014MicrosoftCC}&AP $\uparrow$&0.453&0.427&0.407&\textbf{0.150}\\
\hline
\makecell[l]{Skeleton \\detection}&\makecell[l]{ProMask-\\VGG16 \cite{Bai2020ProMaskPM}}&SK-LARGE \cite{shen2017deepskeleton}&F-measure $\uparrow$&0.746 & 0.739 & \textbf{0.774} &\textbf{0.517}\\
\hline
\end{tabular}}
\end{adjustbox}
\end{table*}

\textbf{Analysis:} The simple wave can effectively attack classification, skeleton detection, semantic segmentation, and object detection models, which need to implicitly or explicitly locate objects in space. It is because CNN forms the receptive field of wave patterns for visual spatial coding when learning these tasks, so it will be so sensitive to the distributed noise of waves. 
When the input wave noise meets with the receptive field of waves, the wave resonance may occur at a specific wavelength and orientation, which will exaggeratedly amplify the noise to destroy the original predicting result. 
Conversely, when the input wave noise has a specific relationship with the spatial representation wave in the model, the wave noise will enhance the spatial representation ability of the model.

\textbf{The value of discovery:}
Our discovery of this wave pattern can explain the reason why the CNN model is so sensitive to the wavelength and orientation of the wave in computer vision tasks involving spatial location. 
Our discovery reveals that the nature of CNN encoding visual space possesses wave patterns to a certain extent. CNN is no longer a black box in terms of visual spatial encoding, it is interpretable.

\textbf{Why does this wave pattern appear?}
In the training process of SpaceNet, after training some epochs, the accuracy of spatial position prediction reaches 0.70, and slight wave patterns appear. When the accuracy of spatial position prediction reaches 0.90, significant wave patterns emerge.

Speculation: The periodicity of the wave can serve as a spatial metric, so this wave spatial coding pattern may be an optimal evolution selection. Similarly, animals searching target in a given space for some time, spatial periodic band cells and grid cells will emerge in the brain cortex.
This formation principle needs to be further studied.

\textbf{Potential value: }
The human visual system can establish a coordinate frame. When people recognize objects, the coordinate frame is involved in the recognition process. 
Until now, it's hard to see something like a "coordinate frame" in artificial neural networks.

In this work, the diffraction and interference patterns exhibit in spatial activity maps of trained models, which indicate that visual space representation possesses the wave nature in artificial neural networks. The wavelength has spatial extension. Our work reveals that the periodicity property of waves provides an interpretable space metric. Waves drawn from two orthogonal directions may play a spatial coordinate frame in artificial neural networks. 
This is a potential application value of our findings.

\textbf{Limitation: }
More work is required to thoroughly understand and apply this underlying mechanism of visual space representation.

\section{Conclusion}
In this paper, we introduce a self-supervised convolutional neural network to perform visual space location, leading to the emergence of single-slit diffraction and double-slit interference patterns of waves. 
We discover wave patterns in CNN learning visual space representation. Our findings reveal the nature of CNN encoding visual space to a certain extent. CNN is no longer a black box in terms of visual spatial encoding, it is interpretable. 
Our findings will inspire researchers to explore the robustness of CNN from a new perspective of the wave.

\begin{ack}
This work was supported by the National Natural Science Foundation of China (No. 61802297) and China Postdoctoral Science Foundation(No. 2021M702598).
\end{ack}

\bibliographystyle{IEEEtran}
\bibliography{scibib}

\end{document}